\documentclass{llncs}
\usepackage{graphicx} 

\title{Maintaining and Managing Road Quality:Using MLP and DNN}
\author{Makgotso Jacqueline Maotwana}
\date{24 May 2024}
\institute{ University of Johannesburg, Corner Kingsway and University Road, Auckland Park,Johannesburg, South Africa }
\begin{document}

\maketitle
\begin{abstract}
Poor roads are a major issue for cars, drivers, and pedestrians since they are a major cause of vehicle damage and can occasionally be quite dangerous for both groups of people (pedestrians and drivers), this makes road surface condition monitoring systems essential for traffic safety, reducing accident rates ad also protecting vehicles from getting damaged.  The primary objective is to develop and evaluate machine learning models that can accurately classify road conditions into four categories: good, satisfactory, poor, and very poor, using a Kaggle dataset of road images. 
To address this, we implemented a variety of machine learning approaches. Firstly, a baseline model was created using a Multilayer Perceptron (MLP) implemented from scratch. Secondly, a more sophisticated Deep Neural Network (DNN) was constructed using Keras. Additionally, we developed a Logistic Regression model from scratch to compare performance. Finally, a wide model incorporating extensive feature engineering was built using the K-Nearest Neighbors (KNN) algorithm with sklearn.The study compared different models for image-based road quality assessment. Deep learning models, the DNN with Keras achieved the best accuracy, while the baseline MLP provided a solid foundation. The Logistic Regression although it is simpler, but it provided interpretability and insights into important features. The KNN model, with the help of feature engineering, achieved the best results. The research shows that machine learning can automate road condition monitoring, saving time and money on maintenance. The next step is to improve these models and test them in real cities, which will make our cities better managed and safer.

\keywords{  Multilayer Perceptron \and Deep Neural Network \and  K-Nearest Neighbors \and Logistic Regression }
\end{abstract}

\section{Introduction}

One of the main measures of road quality (safety or dangerous road) is the state of the road surface. Road anomalies, including cracks, potholes, patches, and tiny surface flaws, can be used to assess the condition of the road surface. Thus, the key to monitoring road quality is identifying potholes, roughness levels, and speed bumps (also known as road bumps). Poor roads are a major issue for cars, drivers, and pedestrians since they are a major cause of vehicle damage and can occasionally be quite dangerous for both groups of people (pedestrians and drivers) . Road surface condition monitoring systems are therefore crucial tools for enhancing traffic safety, lowering accident rates, and shielding cars from harm caused by poor roads \cite{Mohamed}.
\hfill\break
A large percentage of pavements in developed countries are old, and some have road networks that have deteriorated as a result of harsh weather. Meanwhile, because of a sharp rise in the number of cars and frequent use, developing countries experience pavement problems \cite{sensor} 
One can carry out road damage repair and inspection manually, semi-automatic, or completely automated. In order to record and assess the physical state of the road—which is dangerous—experts typically do the manual procedure by walking or driving slowly over the road. Nevertheless, even when the expert does this measurement directly, it's likely that the two experts' evaluations are not fair. A moving vehicle has been employed for data collecting in the semi-automatic method, although damage identification is not done directly. At the same time, the completely automated system uses cars equipped with advanced sensors to gather data and uses image processing techniques to identify the vehicles. Unfortunately all three of these strategies need money, time, and effort \cite{deepLearning}. 
\hfill\break
Recent developments in machine learning (ML) and artificial intelligence (AI) have set up opportunities for improving smart city systems, particularly the transport system. This include boosting overall road safety, facilitating more effective maintenance, and upgrading road condition monitoring systems. These systems' combination of AI and ML algorithms makes it possible to analyse vast amounts of data that have been gathered from several sources, including sensors, cameras, and mobile devices \cite{RoadQuality}. 
Deep learning (DL) and machine learning (ML) are two of these techniques that may be used to train classifiers that discover relationships and trends between input characteristics and class labels. Decision tree, random forest, Support Vector Machine (SVM), and k-Nearest Neighbours (kNN) are a few examples of machine learning algorithms. Road surface irregularities may be found in the machine learning area by manually extracting traits from sensor data. DL techniques, on the other hand, are becoming a popular option for road surface monitoring due to their ability to automatically extract traits from the vibration sensor data. Convolutional neural networks (CNN), long short-term memory (LSTM), and gated recurrent units (GRU) are a few examples of DL techniques \cite{HybridDeep}. 
 This study explores the implementation of the Multilayer perceptron (MLP) and a Deep Neural Network (DNN) with a CNN architecture to classify road quality. It uses the MLP as the baseline model and uses the DNN as the state-of-the-art. This study investigates whether DNN is better at classifying road quality compared to an MLP.  The results of this study may lead to more effective and flexible methods of evaluating road quality, which would be helpful for managing and maintaining roads.  The paper begins with an Introduction that highlights the importance of road quality assessment and the limitations of traditional methods. Then the Related Work section reviews previous research on road quality classification using these techniques and other different techniques to solve this problem. In the Methodology section, the study talks about the used dataset, preprocessing steps, and the architectures and hyperparameters of both the MLP and DNN models. The Results section shows the performance metrics of both models which compares their accuracies and losses. The paper begins with an Introduction that highlights the importance of road quality assessment and the limitations of traditional methods. Then the Related Work section reviews previous research on road quality classification using these techniques and other different techniques to solve this problem. In the Methodology section, the study talks about the used dataset, preprocessing steps, and the architectures and hyperparameters of both the MLP and DNN models. The Results section shows the performance metrics of both models which compares their accuracies and losses.

Which is followed by the Discussion section which analyses the findings, discussing the limitations of the MLP, the improvements seen with the DNN, and the effects of data augmentation and early stopping. 

\section{Related Work}

Li and other authors introduced a novel approach for automating pavement crack classification through deep Convolutional Neural Networks (CNNs) by applying it to 3D pavement images. They used four CNN architectures, denoted as CNN-1 to CNN-4, which were trained and evaluated on a dataset with over 28,000 image patches representing five types of cracks (non-crack, transverse crack, longitudinal crack, block crack, and alligator crack).For each network, the authors use a similar architecture which consist of convolutional layers that have different filter sizes, maxpooling layers and fully connected layers for classification. CNN-1 uses 3x3 filters, CNN-2 uses 5x5 filters, CNN-3 uses 7x7 filters, and CNN-4 uses 9x9 filters. 
Despite the CNNs having different architectures, they all had a high accuracy level, with CNN-3 exceeding 96\%, which was the highest overall accuracy, but the other CNNs also achieved a high accuracy level, which was above 94\% ,thus demonstrating the effectiveness of using CNNs for automatically classifying cracks.\cite{Automatic-classification-of-pavement-crack-using-deep-convolutional-neural-network}. Maeda and the other authors \cite{Road-damage-detection-and-classification-using} used a dataset which was obtained by taking pictures of the roads using a smartphone which was mounted on a car dashboard. 
The dataset has 163,664 images which is categorised into 8 different classes of road damages which were collected from seven municipalities in Japan. 
In their experiments, they trained and evaluated two state-of-the-art object detection models, namely SSD Inception V2 and SSD MobileNet and they used the dataset which they have collected. These models take advantage deep learning techniques to identify and classify road damages within the captured images. 
The hyperparameter settings were detailed for each model which was an initial learning rate of 0.002 and an image size of 300 x 300 pixels for the SSD Inception V2, while SSD MobileNet had an initial learning rate of 0.003 and the same image size. Training and evaluation were performed on a dataset split into training (7,240 images) and evaluation (1,813 images) sets. The authors used data augmentation techniques, like horizontal image flipping, were applied during training to enhance model robustness. The evaluation results demonstrated the effectiveness of SSD MobileNet, which showed good performance in terms of recall, precision, and inference speed compared to SSD Inception V2 \cite{Road-damage-detection-and-classification-using}.

Wang and his other co-authors particpated in the IEE road damage detecion challenge,they introduced a method for road damage detection and classification by using deep learning methods. They proposed a model based on state-of-the-art object detection algorithms like Faster-RCNN and SSD, and they took advantage of the pretrained CNN models like VGG and ResNet which could help in extracting features of images. 
For Faster-RCNN, they used an input size of 600x600 pixels and training was stopped at 60 epochs which had an initial learning rate of 1e-4, and then reduced to one tenth for every 25 epoch. On the other hand, the input size of SSD is originally 1024x1024 and then gradually increased to 1078x1078 after the first 13 epochs, they then set the learning rate for the SSD to 1e-3 and then later reduced it to one-tenth at 11 and 14 epochs. They found that picking at these hyperparameters was highly effective for road damage detection and classification. 
Wang and his other co-authors used the ensemble techniques in order to improve the detection accuracy and they managed to achieve an F1-score of 0.659 which was the best compared to the other teams who had participated in the IEE challenge. This showed that their method was the best for road damage detection and classification\cite{Deep-Proposal-and-Detection-Networks-for-Road-Damage-Detection-and-Classification}.
In this study \cite{A-deep-learning-approach-for-fast-detection-and}, Jiang and other co-authors proposed two improved object detection algorithms, Fast-YOLO and MobileNetV3-SSD, for concrete damage detection based on modifications to existing network architectures. The authors made changes to the YOLOv3 and MobileNetV3 structures in order to improve them. 
For Fast-YOLO, the input image is initially adjusted to 416x416 pixels, then later on compressed to 208x208 pixels, leading to a final feature map size of 208x208x32. Similarly, in MobileNetV3-SSD, the input image is resized to 300x300 pixels before being processed by the network. These adjustments that they made contributed to the detection accuracy of concrete damage types improving, as evidenced by performance metrics such as average precision, which ranges from 54.74\% to 64.81\% for MobileNetV3-SSD models with different scaling ratios. 
During training process, they used the 10-fold cross-verification training strategy and transfer learning method. To optimize the model, they used the Adam optimizer which had initial learning rate set to 0.001 \cite{A-deep-learning-approach-for-fast-detection-and}.

\section{My Idea}
Heavy traffic, bad weather, aging infrastructure and poor maintenance these are the things which could result in damage to the road as time goes by. In order for people to have a more comfortable and safe driving experience, the road must be in a good condition. Although this is something that we want, but sometimes these problems on the road are not found quickly and this could lead to bigger issues. 
This is a big problem in many countries, where some roads are simply just old and they are worn out, while others get damaged by things like extreme weather or too many cars. It could take a lot of time and money to manually check the roads, which is why researchers and engineers are working on more efficient techniques for monitoring road conditions. They want to make sure roads stay safe and strong without costing too much. So, they are developing new methods to check roads regularly and fix any problems before they get worse.
They want to make sure roads are safe and strong enough without costing too much. So, in order to solve this problem, we are developing a new method to check roads on a regular basis and fix any problems before they get worse \cite{Road-condition-monitoring-using-smart-sensing-and-artificial-intelligence}. Our goal is to develop machine learning and deep learning models which will be able to classify the road damages.
\hfill \break
To implement this approach, we train our models using the following :
\subsubsection{Multilayer Perceptron(MLP)}
\hfill\break
We are going implement the baseline Multilayer Perceptron from scratch. The multilayer perceptron (MLP) is a basic type of deep neural network which is known because of its multiple hidden layers which was designed to understand complex patterns in the training data. It is also referred to as the deep feedforward neural network (DFN) \cite{The-multilayer-perceptron-(mlp)}. Which is trained and evaluated using the road damage classification dataset and be able to accurately classify the road conditions as 'Good','Poor', 'Satisfactory' and 'Very Poor'. 

\subsubsection{Deep Neutral Network}
\hfill\break
Deep neural networks (DNN) are commonly used in automatic classification studies due to their ability to adapt and efficiency \cite{Compact-visualization-of-dnn-classification-performances}. To tackle the problem of road damage classification, we use the  Convolutional Neural Networks(CNNs). CNNs are a common type of deep neural network and they made up of multiple CONV layers \cite{Efficient-processing-of-deep-neural-networks}. CNNs are special computer programs inspired by how animals see. They're particularly good at understanding things like images. CNNs automatically learn to identify important features that are in an image, this goes from identifying simple shapes and lines to more complex patterns. They do this through different layers that work together like building blocks, these layers are called convolutional, pooling, and fully connected layers. A convolution layer is the core part of CNN architecture. It is responsible for extracting the features that are in the image. What it does is combine linear and nonlinear operations, such as convolution and activation functions so that it can be able to identify important features in the input data. \cite{Convolutional-neural-networks:-an-overview-and-application-in-radiology}. The pooling layer comes after the convolution layer in a CNN. Its job is to shrink the size of the data, called feature maps. There are two main ways to do this pooling through average pooling and max pooling which are the two most used forms of pooling operations\cite{Deep-learning-for-remote-sensing-image-classification}.
After the convolutional and pooling layers have done their job of extracting and summarising the features that are in an image, the data gets is flattened into a one-dimensional vector. CNN needs to make sense of the features that have been extracted and summarised. This is where fully connected layers come in, the vector is then fed into one or more fully connected layers also known as dense layers. Unlike the previous layers, dense layers have full connections between every neuron in the previous layer and every neuron in the current layer. These dense layers act as a classifier, working together to analyse all the extracted features collectively. A non-linear activation function, such as ReLU, is usually applied after each dense layer. The final fully connected layer has the same number of outputs as the number of classes in the classification task. This layer assigns a probability to each class, which indicates how likely it is that the input falls into one of those classifications\cite{Convolutional-neural-networks:-an-overview-and-application-in-radiology}. So in the context of road damage classification ,CNNs can be used to identify and categorise different types of road damage from the dataset. CNNs will have the ability learn to differentiate between various types of road conditions such as good, poor, satisfactory and very poor. We are also going to take advantage of Tensorflow with Keras which will simplify the implementation of CNNs. 

\subsection{Method}
\subsubsection{Data Collection}
\hfill\break
For the experiment, we collected dataset from Kaggle named Road Quality Classification. Which is divided into 4 categories, these categories are based on the condition of the road  : “Good”, ”Poor”, “Satisfactory” and “ Very Poor”.  Which are shown as follows. 
\begin{figure}
    \centering

    \begin{minipage}{0.49 \textwidth}
    \centering
    \includegraphics[width=0.5\linewidth]{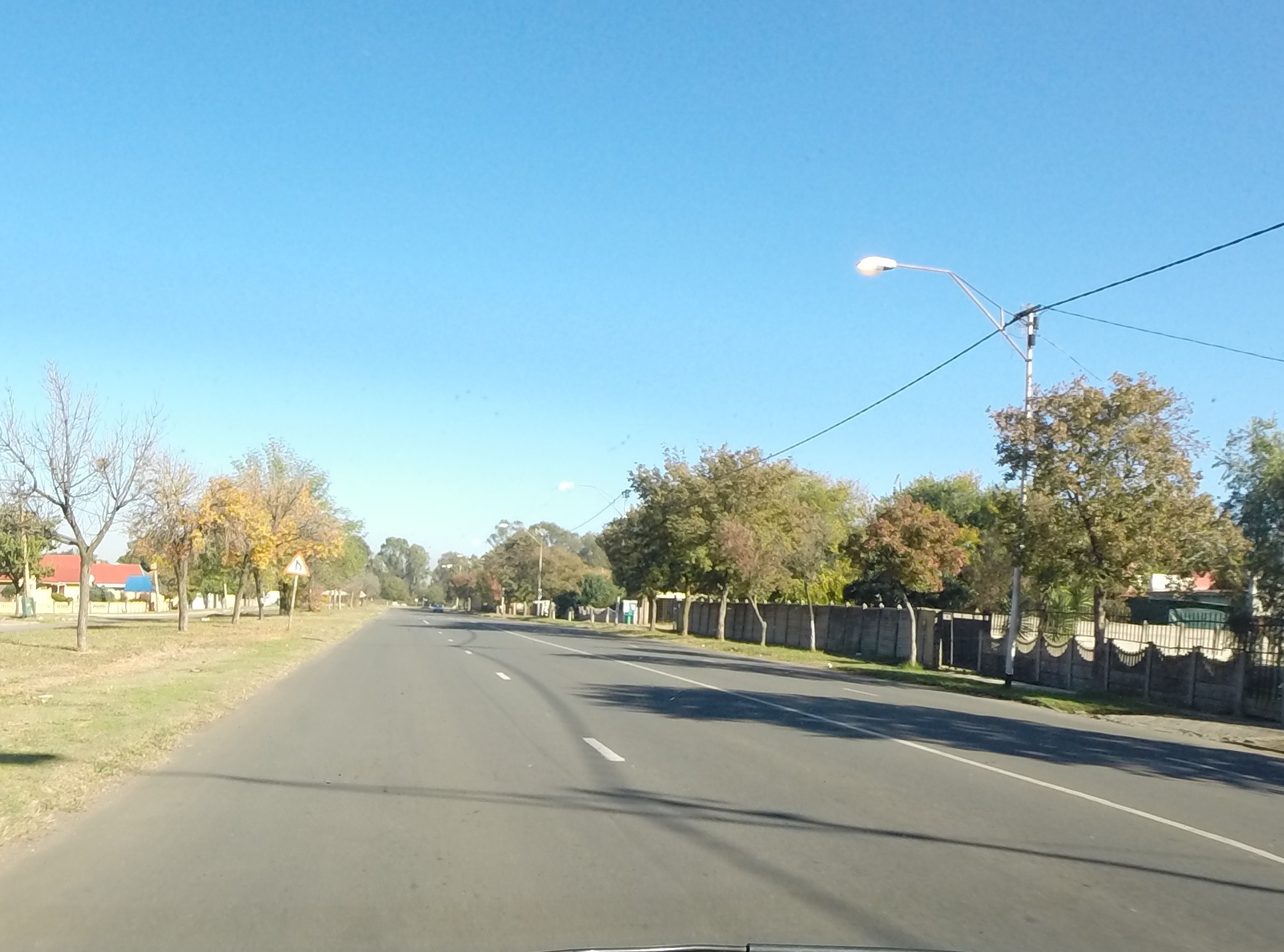}
    \caption{Good}
    \label{fig:enter-label}
\end{minipage}
\begin{minipage}{0.49\textwidth}
 \centering
    \includegraphics[width=0.5\linewidth]{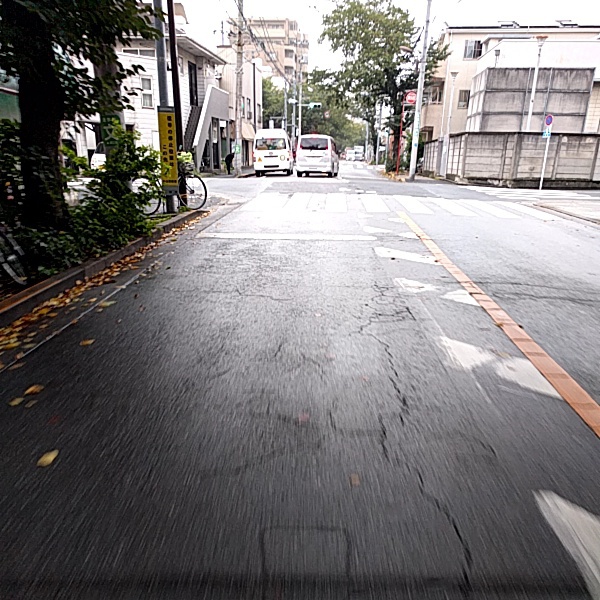}
    \caption{Poor}
    \label{fig:enter-label}
\end{minipage}
    \begin{minipage}{0.49\textwidth}
        
            \centering
             \includegraphics[width=0.5\linewidth]{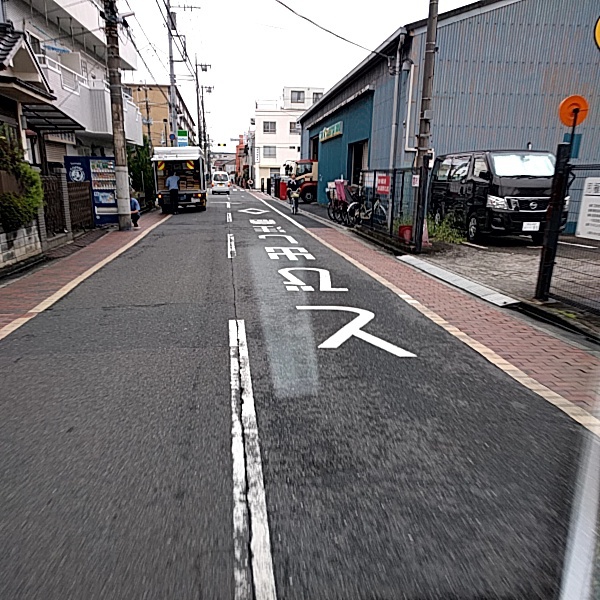}
            \caption{Satisfactory }
            \label{fig:enter-label}
        
    \end{minipage}
    \begin{minipage}{0.49\textwidth}
        
            \centering
            \includegraphics[width=0.5\linewidth]{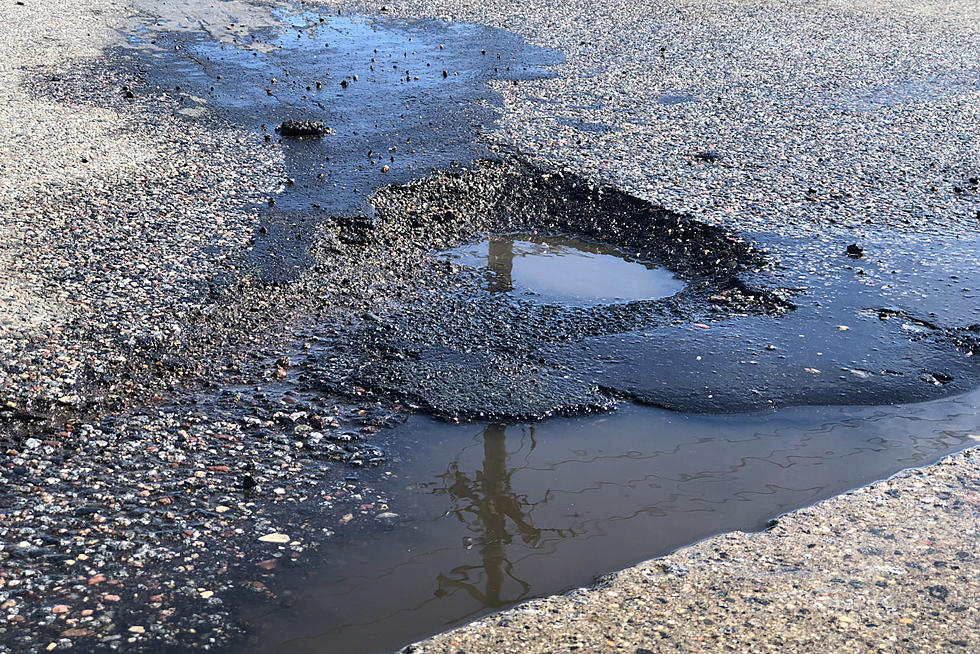}
            \caption{Very Poor }
            \label{fig:enter-label}
        
    \end{minipage}
\end{figure}
\hfil \break

\subsubsection{Preproceess}
\hfill\break
We  process the image data set for training a model in order to transfer I the road conditions. 
So the images are resized to a fixed size of 224. 
The images are then normalised by dividing each pixel value by 255 so that they could range from 0-1.
The data is then split into training and testing sets.
We then use the one hot encoded method in order to convert the data into a binary matrix representation.
We use image data generator to perform data augmentation, which does different transformation to the data such as, horizontal and vertical flipping, shearing, zooming and shifting. By doing this transformations, these transformations are going to help increase the diversity of the training data and improve the models generalisation ability.

\subsubsection{Baseline model:MLP from scratch } 
\hfill\break
Our model is the multi layer perceptron MLP which was implemented from scratch it has an input layer with 150528 neurons for a 224x224x3 image,then a hidden layer which consists of 500 neurons and it also has an output layer which has 4 neurons since we have 4 classes . We train the MLP which follows a sequence of forward backward propagation and then it updates the weights and the Biases.  During the forward propagation we used ReLU activation Function, which was computed in each layer the ReLU combines the inputs with weights and the Biases. This finally leads to the output layer where a softmax function is used In order to compute the probabilities of the different categories.  Backward propagation then computes the gradients of follows function and then it updates the weights and the biases.  This process happens 2000 times with a batch size of 100 and the learning rate is set to 0.001 until the loss function is minimised.

\subsubsection{Deep Neural Network(DNN)}
\hfill\break
We built the deep neutral network using a popular tool that is called the karas which uses convolutional neural Networks CNN. The model starts with a convolutional layer that has 32 filters of size 5 x 5 pixels uses a ReLU activation function. Each convolutional layer is followed by a Max pooling Leah which has a pool size of 2x2 pixels. After the last makes pulling layer a flattening layer is then added so that it can convert the 3D feature maps into a 1D vector and then passed through a dense layer that has 256 neurons and a ReLU activation function .we then add a dropout layer with a dropout rate of 0.3 so that we can prevent overfitting before the final dense layer, which has a number of neurons of four and uses a soft mix function to produce class probabilities. We train the model which uses the Adam optimiser that has a learning rate of 0.01 for 25 epochs, on images of size 224x224 pixels with a 3 colour channels. We then use early stopping when we are training the model to monitor validation loss, the patience is set to 5 which means that the training will stop if the validation loss does not improve for 5 consecutive epochs and it has a batch size of 32.

\section{Results}
\subsubsection{MLP}
\hfill\break
In this study, MLP was used to classify road quality. We trained and evaluated the model using a dataset specifically designed for this classification task. Despite various attempts to optimize the model, the results indicated that both training and test accuracies were relatively low. The results for the MLP accuracies achieved are as follows: 
\begin{itemize}
    \item Training Accuracy: 40.81\%
    \item Test Accuracy: 40.48\%
\end{itemize}
These results suggest that the MLP struggled to effectively learn and generalize from the data provided.

\subsubsection{DNN using Keras}
\hfill\break
DNN with a Convolutional Neural Network architecture was used to classify road quality. The model was trained using the Keras library, and we used different methods to optimize the performance such as data augmentation and early stopping. The model stops at 21/25 epochs and achieved a test accuracy of 89.31\% with a loss of 0.2748 and training accuracy of 93.82\% with a loss of 0.1859. The evaluation metrics indicate that the model performs well on both the training and test data, though there is a noticeable gap between training and test accuracy,meaning there could be some overfitting.

These results show that the model initially showed improvement in both training and validation accuracy and loss, particularly in the first few epochs. The figure \ref{Loss vs Validation Lossl} below shows the training loss and the validation loss. The  training loss graph (blue line) start very high but as time goes by it decreases,it gets better at fitting the data, so the loss naturally goes down . There are some fluctuations, but the overall trend is downward which suggests that the model is effectively learning from the training data.The  validation loss (orange line) also starts high and generally follows a similar downward trend as the training loss, indicating that the model is also improving its performance on the validation data.
However, the validation loss fluctuates more than the training loss and has some noticeable spikes. This could suggest periods of overfitting where the model is learning specific patterns in the training data that do not generalize well to the validation data.
After about 10 epochs, both the training and validation loss appear to be stable, although the validation loss still shows some spikes.
The model seems to have reached a point where further training does not significantly reduce the validation loss, indicating that it might be close to the optimal performance on the validation set. The presence of spikes in the validation loss, while the training loss continues to decrease, suggests that the model may be overfitting the training data. This means that while the model performs well on the training data, it does not generalize as well to unseen validation data.

\begin{figure}
    \centering
    \includegraphics[width=0.85\linewidth]{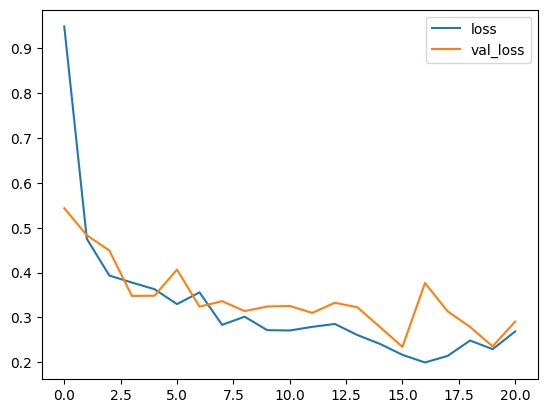}
    \caption{Loss vs Validation Loss}
    \label{Loss vs Validation Lossl}
\end{figure}

\newpage
\section{Discussion}
We did different approaches in order to enhance the performance of the MLP, such as tuning hyperparameters. We tried to make adjustments to the learning rate, number of hidden layers, number of neurons per layer, and the number of training epochs. However, all of these adjustments did not improve the accuracies making the MLP model not to be the ideal solution for the road classification problem. 
This could mean that the MLP might be too simple to capture the underlying patterns of the road quality classification problem. More complex models, such as CNNs which might be better suited for this task. We could also use data augmentation to add noise to our dataset which could help improve the accuracy of the model. We then used DNN model  with an architecture of CNN which showed strong learning because its accuracy went up and its loss went down during training.Making the training data have more noise with the use of data augmentation and early stopping likely improved the model's ability to generalize well to new data and prevent the model from overfitting. 
The model started well but after a while it showed some instability. We can see that with the fluctuations in validation accuracy and loss.  After reaching its best performance at epoch 9, the validation accuracy dropped to 87.28\% at epoch 10, and the validation loss shifted, indicating the model might have learned the data too well and it is now overfitting even though we did early stopping. The reason for overfitting could be issues with the data, like noise or missing examples for certain road types, which could be limiting the model's ability to learn and apply its knowledge to new data basically generalize well. 

\section{Conclusion}
A Multi-Layer Perceptron model applied to road quality classification performed poorly, achieving only around 40\% accuracy on both training and testing data. Tweaking the MLP's hyperparameters did not help improve the results, this means that it might not be the best choice for this task. The use of a DNN with a CNN architecture for road quality classification yielded promising results, with training accuracy reaching 93.82\% and the testing  accuracy at 89.31\%. The application of data augmentation and early stopping proved beneficial in enhancing model robustness and preventing overfitting. This means that the DNN is the best model which we could use to help us classify road quality even though at some point it also struggles to generalize well with the new data but with good quality data it could be more but we could use it to properly manage and maintain the road. 
\hfill\break
Even though the CNN model performed well, but there is always room for improvement in future research. The model's performance could be improved by using more and better dataset. This means collecting a wider variety of road quality examples and cleaning the data to remove errors and unnecessary information. We could try more advanced model designs, like combining CNNs with RNNs to see if they can learn even more complex patterns from the data.
The next step could be building a real-time system for road quality assessment running on phones and roadside equipment, helping transportation departments and road agencies with their work.

\newpage
\bibliographystyle{splncs03_unsrt}
\bibliography{bibliography}
\end{document}